\documentclass[10pt,twocolumn,letterpaper]{article}
\pdfoutput=1

\usepackage{cvpr}
\usepackage{times}
\usepackage{epsfig}
\usepackage{graphicx}
\usepackage{amsmath}
\usepackage{amssymb}
\usepackage{authblk}
\usepackage{booktabs,multirow}
\usepackage[ruled]{algorithm2e}
\usepackage[inline]{enumitem}
\usepackage{cite}

\usepackage[breaklinks=true,bookmarks=false,hidelinks]{hyperref}

\cvprfinalcopy % *** Uncomment this line for the final submission

\def\cvprPaperID{****} % *** Enter the CVPR Paper ID here

\DeclareMathOperator*{\E}{\mathbb{E}}

\graphicspath{{figs/}}

\begin{document}

%%%%%%%%% TITLE
\title{
Large-Scale Distributed Second-Order Optimization Using Kronecker-Factored Approximate Curvature for Deep Convolutional Neural Networks
}

\makeatletter
\renewcommand{\AB@affilsep}{\quad\protect\Affilfont}
\renewcommand\Authands{\quad}
\renewcommand\Authsep{\quad}
\makeatother

\author[1]{Kazuki Osawa}
\author[1]{Yohei Tsuji}
\author[1]{Yuichiro Ueno}
\author[3]{Akira Naruse}
\author[2]{Rio Yokota}
\author[4,1]{Satoshi Matsuoka}
\affil[1]{School of Computing, Tokyo Institute of Technology}
\affil[2]{Global Scientific Information and Computing Center, Tokyo Institute of Technology}
\affil[3]{NVIDIA}
\affil[4]{RIKEN Center for Computational Science}
\affil[ ]{\tt\small \{oosawa.k.ad, tsuji.y.ae, ueno.y.ai\}@m.titech.ac.jp}
\affil[ ]{\tt\small anaruse@nvidia.com, rioyokota@gsic.titech.ac.jp, matsu@is.titech.ac.jp}

\maketitle
%\thispagestyle{empty}

%%%%%%%%% ABSTRACT
\begin{abstract}
    Large-scale distributed training of deep neural networks suffer from the generalization gap caused by the increase in the effective mini-batch size. Previous approaches try to solve this problem by varying the learning rate and batch size over epochs and layers, or some \textit{ad hoc} modification of the batch normalization. We propose an alternative approach using a second-order optimization method that shows similar generalization capability to first-order methods, but converges faster and can handle larger mini-batches. To test our method on a benchmark where highly optimized first-order methods are available as references, we train ResNet-50 on ImageNet. We converged to 75\% Top-1 validation accuracy in 35 epochs for mini-batch sizes under 16,384, and achieved 75\% even with a mini-batch size of 131,072, which took only 978 iterations.
\end{abstract}

%%%%%%%%% BODY TEXT
\section{Introduction}
As the size of deep neural network models and the data which they are trained on continues to increase rapidly,
the demand for distributed parallel computing is increasing.
A common approach for achieving distributed parallelism in deep learning is to use the data-parallel approach,
where the data is distributed across different processes while the model is replicated across them.
When the mini-batch size per process is kept constant to increase the ratio of computation over communication,
the effective mini-batch size over the entire system grows proportional to the number of processes.

When the mini-batch size is increased beyond a certain point, the validation accuracy starts to degrade.
This generalization gap caused by large mini-batch sizes have been studied extensively
for various models and datasets \cite{shallue2018}.
Hoffer \etal attribute this generalization gap to the limited number of updates, and suggest to train longer \cite{hoffer2018}.
This has lead to strategies such as scaling the learning rate proportional to the mini-batch size,
while using the first few epochs to gradually warmup the learning rate \cite{smith2017}.
Such methods have enabled the training for mini-batch sizes of 8K,
where ImageNet \cite{deng2012} with ResNet-50 \cite{he2015a} could be trained for 90 epochs
to achieve 76.3\% top-1 validation accuracy in 60 minutes \cite{goyal2017}.
%on 256 Tesla P100 GPUs
Combining this learning rate scaling with other techniques such as RMSprop warm-up,
batch normalization without moving averages, and a slow-start learning rate schedule,
Akiba \etal were able to train the same dataset and model with a mini-batch size of 32K
to achieve 74.9\% accuracy in 15 minutes \cite{akiba2017}.
%to achieve 74.9\% Top-1 validation accuracy in 15 minutes on 1024 Tesla P100 GPUs \cite{akiba2017}.
%Spending longer number of trials for the hyperparameter tuning for modified batch normalization,
%aggressive learning rate schedules, warm-up strategies, weight decay improvements,
%and using collapsed ensembles has shown to yield Top-1 validation accuracies of 76.6\% for 8K,
%76.26\% for 16K, and 75.31\% for 32K batch sizes \cite{codreanu2017}.

More complex approaches for manipulating the learning rate were proposed, such as LARS \cite{you2017},
where a different learning rate is used for each layer by normalizing them with
the ratio between the layer-wise norms of the weights and gradients.
%This enabled the training of the same dataset and model with a mini-batch size of 32k
This enabled the training with a mini-batch size of 32K
without the use of \textit{ad hoc} modifications, which achieved 74.9\% accuracy
%in 14 minutes (64 epochs) on 2048 Xeon Phi (Knights Landing) co-processors \cite{you2017}.
in 14 minutes (64 epochs) \cite{you2017}.
It has been reported that combining LARS with counter intuitive modifications to the batch normalization,
can yield 75.8\% accuracy even for a mini-batch size of 65K \cite{jia2018}.
%Implementing this with mixed precision allows Jia \etal to train ImageNet-1K with ResNet-50
%in 6.6 minutes (90 epochs) on 2048 Tesla P40 GPUs.

The use of small batch sizes to encourage rapid convergence in early epochs,
and then progressively increasing the batch size is yet another successful approach \cite{smith2017,devarakonda2017}.
Using such an adaptive batch size method, Mikami \etal were able to train 122 seconds
 with an accuracy of 75.3\%.
The hierarchical synchronization of mini-batches have also been proposed \cite{lin2018},
but such methods have not been tested at scale to the extent of the authors' knowledge.

In the present work, we take a more mathematically rigorous approach to tackle the large mini-batch problem,
by using second-order optimization methods.
We focus on the fact that for large mini-batch training,
each mini-batch becomes more statistically stable and falls into the realm
where second-order optimization methods may show some advantage.
Another unique aspect of our approach is the accuracy at which we can approximate
the Hessian when compared to other second-order methods.
Unlike methods that use very crude approximations of the Hessian, such as the TONGA \cite{leroux2008},
Hessian free methods \cite{martens2010},
we adopt the Kronecker-Factored Approximate Curvature (K-FAC) method \cite{martens2015}.
The two main characteristics of K-FAC are that it converges faster than first-order
stochastic gradient descent (SGD) methods,
and that it can tolerate relatively large mini-batch sizes without any \textit{ad hoc} modifications.
K-FAC has been successfully applied to convolutional neural networks \cite{grosse2016},
distributed memory training of ImageNet \cite{ba2017}, recurrent neural networks \cite{martens2018},
Bayesian deep learning \cite{zhang2017}, and reinforcement learning \cite{wu2017}.

Our contributions are:
\begin{itemize}
    \item
        We implement a distributed K-FAC optimizer using a synchronous all-worker scheme.
        We used half precision floating point numbers for both computation
        and exploited the symmetry of the Kronecker factor to reduce the overhead.
    \item
    	We were able to show for the first time that second-order optimization methods can
	achieve similar generalization capability compared to highly optimized SGD,
	by training ResNet-50 on ImageNet as a benchmark.
	We converged to 75\% top-1 validation accuracy in 35 epochs for mini-batch sizes under 16,384,
    and achieved 75\% even with a mini-batch size of 131,072, which took only 978 iterations (Table~\ref{tab:summary}).
    \item
        We show that we can reduce the frequency of updating the Fisher matrices for K-FAC after a few hundred iterations.
        In doing so, we are able to reduce the overhead of K-FAC.
        We were able to train ResNet-50 on ImageNet in 10 minutes to a top-1 accuracy of 74.9\%
        using 1,024 Tesla V100 GPUs (Table~\ref{tab:compare}).
    \item
    We show that the Fisher matrices for Batch Normalization layers \cite{ioffe2015} can be approximated as diagonal matrices,
        which further reduces the computation and memory consumption.
\end{itemize}

\begin{table}[ht]%{{{
    \begin{center}
        \caption{Training epochs (iterations) and top-1 single-crop validation accuracy of ResNet-50 for ImageNet with K-FAC.}
        \label{tab:summary}
        \begin{tabular}{c c c c}
            \toprule
            \bf Mini-batch size & \bf Epoch & \bf Iteration & \bf Accuracy \\
            \midrule
            4,096 & 35 & 10,948 & 75.1 $\pm$ 0.09\,\% \\
            8,192 & 35 & 5,434 & 75.2 $\pm$ 0.05\,\% \\
            16,384 & 35 & 2,737 & 75.2 $\pm$ 0.03\,\% \\
            32,768 & 45 & 1,760 & 75.3 $\pm$ 0.13\,\% \\
            65,536 & 60 & 1,173 & 75.0 $\pm$ 0.09\,\% \\
            131,072 & 100 & 978 & 75.0 $\pm$ 0.06\,\% \\
            \bottomrule
        \end{tabular}
    \end{center}
\end{table}%}}}

\section{Related work}

\begin{table*}[ht]%{{{
    \begin{center}
        \caption{Training iterations (time) and top-1 single-crop validation accuracy of ResNet-50 for ImageNet reported by related work.}
        \label{tab:compare}
        \small
        \begin{tabular}{c c c c c c c c}
            \toprule
            &\bf Hardware&\bf Software&\bf Mini-batch size&\bf Optimizer&\bf Iteration&\bf Time&\bf Accuracy\\
            \midrule
            Goyal \etal \cite{goyal2017} & Tesla P100 $\times$ 256 & Caffe2 & 8,192 & SGD & 14,076 & 1 hr & 76.3\% \\
            %You \etal \cite{you2017} & Xeon 8160 $\times$ 1600 & Intel Caffe & 16,000 & SGD & 90 & 31 min & 75.3\% \\
            You \etal \cite{you2017} & KNL $\times$ 2048 & Intel Caffe & 32,768 & SGD & 3,519 & 20 min & 75.4\% \\
            Akiba \etal \cite{akiba2017} & Tesla P100 $\times$ 1024 & Chainer & 32,768 & RMSprop/SGD & 3,519 & 15 min & 74.9\% \\
            You \etal \cite{you2017} & KNL $\times$ 2048 & Intel Caffe & 32,768 & SGD & 2,503 & 14 min & 74.9\% \\
            %Jia \etal \cite{jia2018} & Tesla P40 $\times$ 1024 & TensorFlow & 65,536 & SGD & 90 & 8.7 min & 76.2\% \\
            Jia \etal \cite{jia2018} & Tesla P40 $\times$ 2048 & TensorFlow & 65,536 & SGD & 1,800 & 6.6 min & 75.8\% \\
            Ying \etal \cite{ying2018a} & TPU v3 $\times$ 1024 & TensorFlow & 32,768 & SGD & 3,519 & 2.2 min & 76.3\% \\ 
            Mikami \etal \cite{mikami2018} & Tesla V100 $\times$ 3456 & NNL & 55,296 & SGD & 2,086 & 2.0 min & 75.3\% \\
            \midrule
            This work (Sec. \ref{subsec:fisher}) & Tesla V100 $\times$ 1024 & Chainer & 32,768 & K-FAC & 1,760 & 10 min & 74.9\% \\
            This work (Sec. \ref{subsec:training}) & - & Chainer & 131,072 & K-FAC & 978 & - & 75.0\% \\
            \bottomrule
        \end{tabular}
    \end{center}
\end{table*}
%}}}

With respect to large-scale distributed training of deep neural networks,
there have been very few studies that use second-order methods.
At a smaller scale, there have been previous studies that used K-FAC to train ResNet-50 on ImageNet \cite{ba2017}.
However, the SGD they used as reference was not showing state-of-the-art Top-1 validation accuracy
(only around 70\%), so the advantage of K-FAC over SGD that they claim was not obvious from the results.
In the present work, we compare the Top-1 validation accuracy with state-of-the-art SGD methods for large mini-batches
mentioned in the introduction (Table~\ref{tab:compare}).

The previous studies that used K-FAC to train ResNet-50 on ImageNet \cite{ba2017}
also were not considering large mini-batches and were only training with mini-batch size of 512 on 8 GPUs.
In contrast, the present work uses mini-batch sizes up to 131,072, which is equivalent to 32 per GPU on 4096 GPUs,
and we are able to achieve a much higher Top-1 validation accuracy of 75\%.
Note that such large mini-batch sizes can also be achieved by accumulating the gradient over multiple iterations
before updating the parameters, which can mimic the behavior of the execution on many GPUs
without actually running them on many GPUs.

The previous studies using K-FAC also suffered from large overhead of the communication
since they implemented their K-FAC in TensorFlow \cite{martin2015} and used a parameter-server approach.
Since the parameter server requires all workers to
send the gradients and receive the latest model's parameters from the parameter server,
the parameter server becomes a huge communication bottleneck especially at large scale.
Our implementation uses a decentralized approach using MPI/NCCL\footnote{https://developer.nvidia.com/nccl} collective communications among the processes.
Although, software like Horovod\footnote{https://github.com/horovod/horovod} can alleviate the problems with parameter servers,
the decentralized approach has been used in high performance computing for a long time,
and is known to scale to thousands of GPUs without modification.

\section{Distributed K-FAC}

\subsection{Notation and background}
Throughout this paper, we use $\E[\cdot]$ as the mean among the samples in the mini-batch $\{(\mathbf{x},\mathbf{y})\}$, and compute the {\it cross-entropy loss} as
\begin{equation}
    \label{eq:loss}
    \mathcal{L}(\boldsymbol\theta) = \E[-\log p(\mathbf{y}|\mathbf{x};\boldsymbol\theta)]\,.
\end{equation}
where $\mathbf{x}, \mathbf{y}$ are the training input and label (one-hot vector), $p(\mathbf{y}|\mathbf{x};\boldsymbol\theta)$ is the likelihood calculated by the probabilistic model using a deep neural network with the parameters $\boldsymbol\theta\in\mathbb{R}^N$.

\medskip
\noindent
{\bf Update rule of SGD.}
For the standard first-order stochastic gradient descent (SGD), the parameters $\mathbf{w}_\ell\in\mathbb{R}^{N_\ell}$ in the $\ell$-th layer is updated based on the gradient of the loss function:
\begin{align}
    \mathbf{w}_\ell^{(t+1)}
    &\leftarrow
    \mathbf{w}_\ell^{(t)}
    -
    \eta
    \nabla\mathcal{L}_\ell^{(t)}\,.
\end{align}
where $\eta>0$ is the learning rate and $\nabla\mathcal{L}_\ell\in\mathbb{R}^{N_\ell}$ represents the gradient of the loss function for $\mathbf{w}_\ell$.

\medskip
\noindent
{\bf Fisher information matrix.}
The {\it Fisher information matrix} (FIM) of the probabilistic model is estimated by 
\begin{equation}
    \mathbf{F}_{\boldsymbol\theta} = \E[\nabla\log p(\mathbf{y}|\mathbf{x};\boldsymbol\theta)\nabla\log p(\mathbf{y}|\mathbf{x};\boldsymbol\theta)^{\mathrm{T}}]
    \in\mathbb{R}^{N\times N}.
\end{equation}
Strictly speaking, $\mathbf{F}_{\boldsymbol\theta}$ is the {\it empirical} (stochastic version of) FIM \cite{martens2015}, but we refer to this matrix as FIM throughout this paper for the sake of brevity.
In the training of deep neural networks, FIM can be assumed as the curvature matrix in the parameter space \cite{amari1998,martens2015,botev2017}.
\subsection{K-FAC}
Kronecker-Factored Approximate Curvature (K-FAC) \cite{martens2015} is a second-order optimization method for deep neural networks, which is based on an accurate and mathematically rigorous approximation of the FIM.
K-FAC is applied to the training of convolutional neural networks, which minimizes the log likelihood (\eg a classification task with a loss function (\ref{eq:loss})).

For the training of the deep neural network with $L$ layers, K-FAC approximates $\mathbf{F}_{\boldsymbol\theta}$ as a diagonal block matrix: 
\begin{equation}
    \label{eq:blockdiag}
    \mathbf{F}_{\boldsymbol\theta}
    \approx
    {\rm diag}
    \left(
    \mathbf{F}_{1}
    ,\dots,
    \mathbf{F}_{\ell}
    ,\dots,
    \mathbf{F}_{L}
    \right)
    \,.
\end{equation}
The diagonal block $\mathbf{F}_\ell\in\mathbb{R}^{N_{\ell}\times N_\ell}$ represents the FIM for the $\ell$ th layer of the deep neural network with weights $\mathbf{w}_\ell\in\mathbb{R}^{N_l}$ ($\ell=1,\dots,L$).
Each diagonal block matrix $\mathbf{F}_{\ell}$ is approximated as a Kronecker product:
\begin{equation}
    \label{eq:kf}
    \mathbf{F}_{\ell}\approx \mathbf{G}_{\ell}\otimes \mathbf{A}_{\ell-1}\,(\ell=1,\dots,L )
    \,.
\end{equation}
This is called {\it Kronecker factorization} and 
$\mathbf{G}_{\ell},\mathbf{A}_{\ell-1}$ are called {\it Kronecker factors}.
$\mathbf{G}_{\ell}$ is computed from the gradient of the loss with regard to the output of the $\ell$ th layer, and $\mathbf{A}_{\ell-1}$ is computed from the activation of the $\ell-1$ th layer (the input of $\ell$ th layer) \cite{martens2015,grosse2016}.

The inverse of a Kronecker product is approximated by the Kronecker product of the inverse of each Kronecker factor.
\begin{equation}
    \label{eq:inv_fim}
    \mathbf{F}_{\ell}^{-1}
    \approx
    \left(
        \mathbf{G}_{\ell}\otimes\mathbf{A}_{\ell-1}
    \right)^{-1}
    =
    \mathbf{G}_{\ell}^{-1}
    \otimes
    \mathbf{A}_{\ell-1}^{-1}
    \,.
\end{equation}

\medskip
\noindent
{\bf Update rule of K-FAC.} The parameters $\mathbf{w}_\ell$ in the $\ell$ th layer is updated as follows:
\begin{align}
    \mathcal{G}_\ell^{(t)}
    &=
    \left(
        {\mathbf{G}_\ell^{(t)}}^{-1}
        \otimes
        {\mathbf{A}_{\ell-1}^{(t)}}^{-1}
    \right)
    \nabla\mathcal{L}_\ell^{(t)}\,,
    \\
    \mathbf{w}_\ell^{(t+1)}
    &\leftarrow
    \mathbf{w}_\ell^{(t)}
    -
    \eta
    \mathcal{G}_\ell^{(t)}\,.
\end{align}
where $\mathcal{G}_\ell$ is the {\it preconditioned gradient}.

\begin{figure*}
  \centering
  \includegraphics[width={0.85\textwidth}]{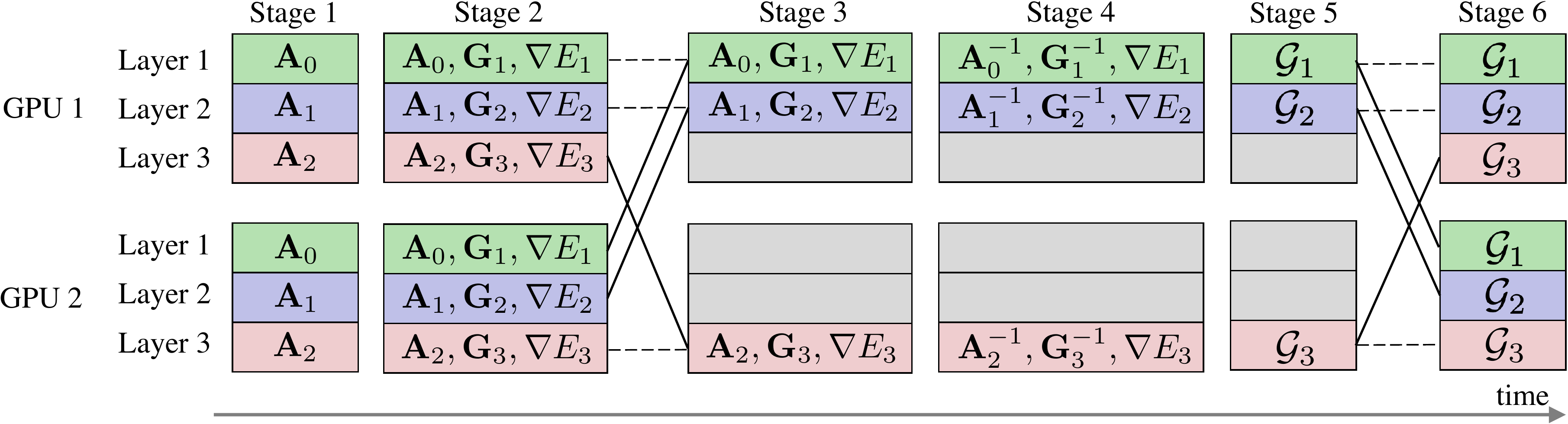}
  \caption{\label{fig:dist_kfac}
    Overview of our distributed K-FAC design.
    There are two processes (GPUs) training on three layers model.
  }
\end{figure*}

\subsection{Our design}
Due to the extra calculation of the inverse FIM, K-FAC has considerable overhead compared to SGD.
We designed a distributed parallelization scheme
so that this overhead decreases as the number of processes is increased.
Furthermore, we introduce a relaxation technique to reduce
the computation of the FIM, which is explained in Section \ref{subsec:fisher}.
In doing so, we were able to reduce the overhead of K-FAC
to almost a negligible amount.

% TODO: remove comment out for rebuttal
%For details of the implementation and the computational optimizations,
%we refer the reader to the published code at (our code will be publicly available if this paper is accepted).

%We put several assumptions on our design to simplify the explanation.
%We use all-worker scheme for distribution, which differs from parameter-server
%scheme in how they communicate and how they locate the model.
%We only consider synchronous parameter update and not asynchronous update. 
%The asynchronous update usually have a potential to speedup the training,
%however the staleness and the non-deterministic behavior introduced by its asynchrony
%let the training and the analysis more difficult \cite{45187}.
%These assumptions are common in the large-scale distributed deep neural
%network training area \cite{akiba2017,goyal2017,jia2018,mikami2018}.

Figure \ref{fig:dist_kfac} shows the overview of our design,
which shows a single iteration of the training.
We use the term \textit{stage} to refer to each phase of the computation, which is indicated at the top of the figure.
The variables in each box illustrates what the process computes during that stage,
\eg at stage 1, each process computes the Kronecker factor $\mathbf{A}$ from the activation.

Stage 1 and 2 are the forward pass and backward pass,
in which the Kronecker factors $\mathbf{A}_{\ell - 1}$ and $\mathbf{G}_{\ell}$
are computed, respectively.
Since the first two stages are computed in a data-parallel fashion,
each process computes the Kronecker factors for all layers,
but using different mini-batches.
In order to get the Kronecker factors for the global mini-batch,
we need to average these matrices over all processes.
This is performed by a ReduceScatterV collective communication,
which essentially transitions our approach from data-parallelism to model-parallelism.
This collective is much more efficient than an AllReduce,
and distributes the Kronecker factors to different processes while summing their values.
Stage 3 shows the result after the communication,
where the model is now distributed across the GPUs.

Stage 4 is the matrix inverse computation and stage 5 is the
matrix-matrix multiplication for computing the pre-conditioned gradient $\mathcal{G}_{\ell}$.
These computations are performed in a model-parallel fashion,
where each processes inverts the Kronecker factors 
and multiplies them to the gradient of different layers.
When the number of layers is larger than the number of GPUs,
multiple layers are handled by each GPU, as shown in Figure \ref{fig:dist_kfac}.
If the number of layers is smaller than the number of GPUs,
some layers will be calculated redundantly on multiple GPUs.
This simplifies the implementation, reduces the communication,
and prevents load-imbalance.

Once we obtain the pre-conditioned gradients,
we switch back to data-parallelism by calling an AllGatherV collective.
After stage 6 is finished, all processes can update their parameters using
the pre-conditioned gradients for all layers.
As we will mention in Section \ref{subsec:fisher},
we are able to reduce the amount of communication required
for the Kronecker factors $\mathbf{A}$ and $\mathbf{G}$.
Therefore, the amount of communication is similar to SGD,
where the AllReduce is implemented as a ReduceScatter+AllGather.
Algorithm \ref{alg:distkfac} shows the pseudo code of our distributed
K-FAC design.
\begin{algorithm}[h]
  \caption{\label{alg:distkfac}Distributed K-FAC Optimizer}
  \DontPrintSemicolon  % dont print semicolon after each line
  \While{not converge}{
    \ForEach{$\ell = 1, \cdots, L$} {
      forward $\ell$ and compute $\mathbf{A}_{\ell - 1}$
    }
    \tcp{stage 1 done}
    \ForEach{$\ell = L, \cdots, 1$} {
      backward $\ell$ and compute $\mathbf{G}_{\ell}$
    }
    \tcp{stage 2 done}
    Reduce+ScatterV$\left(\nabla E_{1:L},
    \mathbf{A}_{0:L-1}, \mathbf{G}_{1:L}\right)$\;
    \tcp{stage 3 done}
    \ForPar{$\ell = 1, \cdots, L$} {
      compute $\mathbf{G}^{-1}_{\ell}$, $\mathbf{A}^{-1}_{\ell - 1}$,
      and $\mathcal{G}_{\ell}$
    }
    \tcp{stage 4 and 5 done}
    AllGatherV$\left(\mathcal{G}_{1:L}\right)$\;
    \tcp{stage 6 done}
    update $\boldsymbol{\theta}$ using $\mathcal{G}_{1:L}$
  }
  \Return $\boldsymbol{\theta}$\;
\end{algorithm}

% 2 column

% 1 column
%\begin{figure}
%  \centering
%  \includegraphics[width={0.9\linewidth}]{distkfac.pdf}
%  \caption{\label{fig:dist_kfac}
%    Overview of our distributed K-FAC design.
%  }
%\end{figure}

\subsection{Further acceleration}
Our data-parallel and model-parallel hybrid approach allows us to 
minimize the overhead of K-FAC in a distributed setting.
However, K-FAC still has a large overhead compared to SGD.
There are two hotspots in our distributed K-FAC design.
The first is the construction of Kronecker factors,
which cannot be done in a model-parallel fashion.
The second is the extra communication for distributing these Kronecker factors.
In this section, we discuss how we accelerated these two hotspots
to achieve faster training time.

\medskip
\noindent
\textbf{Mixed-precision computation.}
K-FAC requires the construction of Kronecker factors $\mathbf{G}_{\ell}$ and
$\mathbf{A}_{\ell - 1}$ for all layers $\ell=1,\cdots,L$ in the model.
Since this operation must be done before taking the global average,
it is in the data-parallel stages of our hybrid approach.
Therefore, its computation time does not decrease even when more processes are used,
and becomes relatively heavy compared to the other stages.
To accelerate this computation, we use the Tensor Cores in the NVIDIA Volta Architecture \footnote{https://www.nvidia.com/en-us/data-center/tensorcore/}.
This more than doubles the speed of the calculation for this part.
%One might think, that this low-precision computation affects the overall accuracy
%of the training. In our experiments we cannot find any differences between using
%half-precision floating point computation or not.
%
%\medskip
%\noindent
%\textbf{Mixed-precision communication.}
%%Our design performs two types of collective communication,
%%which are Reduce+ScatterV and AllGatherV.
%We apply half-precision floating point also to communication.
%Recent studies have found that it is possible to use half-precision
%floating point numbers or even less for communication in data-parallel SGD.
%We apply this technique to reduce the data amount communicated between GPUs.
%%Although, ideally we can achieve double performance compared
%%to single-precision, in practice, it requires scaling values to deal with
%%underflow and overflow due to its limited dynamic range.
%%Even with the extra computation we could achieve better performance.
%Using half-precision floating point in computation and communication might
%affect the accuracy. However, in our experiments we cannot detect any changes
%with switching these methods on.

\medskip
\noindent
\textbf{Symmetry-aware communication.}
The Kronecker factors are all symmetric matrices \cite{martens2015, grosse2016},
so we exploit this property to reduce the volume of communication.
To communicate a symmetric matrix of size $N \times N$,
we only need to send the upper triangular matrix of size $N(N + 1)/2$.
%This means for our example, ResNet-50, we can reduce the amount of the communication
%to 391 MiBytes, which results in a 43\% reduction.

\section{Training schemes}
\label{sec:training}
The behavior of K-FAC on large models and datasets has not been studied in length.
Also, there are very few studies that use K-FAC for large mini-batches (over 4K)
using distributed parallelism at scale \cite{ba2017}.
Contrary to SGD, where the hyperparameters have been optimized by many practitioners even for large mini-batches,
there is very little insight on how to tune hyperparameters for K-FAC.
In this section, we have explored some methods, which we call training schemes,
to achieve higher accuracy in our experiments.
In this section, we show those training schemes in our large mini-batch training with K-FAC.

\subsection{Data augmentation}
\label{subsec:data}%{{{
We resize the all images in ImageNet to $256\times256$ ignoring the aspect ratio of original images and compute the mean value ($224\times224$) of the upper left of the resized images.
When reading an image, we randomly crop a $224\times224$ image from it,
randomly flip it horizontally, subtract the mean value, and scale every pixel to $[0, 1]$.

\medskip
\noindent
{\bf Running mixup.} We extend {\it mixup} \cite{zhang2017a,guo2018} to increase its regularization effect.
We synthesize virtual training samples from raw samples and virtual samples from the previous step
(while the original {\it mixup} method synthesizes only from the raw samples):
\begin{align}
    \tilde{\mathbf{x}}^{(t)}&=\lambda\cdot \mathbf{x}^{(t)}+(1-\lambda)\cdot\tilde{\mathbf{x}}^{(t-1)}\,,
    \\
    \tilde{\mathbf{y}}^{(t)}&=\lambda\cdot \mathbf{y}^{(t)}+(1-\lambda)\cdot\tilde{\mathbf{y}}^{(t-1)}\,.
\end{align}
$\mathbf{x}^{(t)},\mathbf{y}^{(t)}$ is a raw input and label (one-hot vector),
and $\tilde{\mathbf{x}}^{(t)},\tilde{\mathbf{y}}^{(t)}$ is a virtual input and label for $t$ th step.
$\lambda$ is sampled from the Beta distribution with the beta function
\begin{equation}
B(\alpha,\beta)
=\int_0^1t^{\alpha-1}(1-t)^{\beta-1}\,\mathrm{d}t\,.
\end{equation}
where we set $\alpha=\beta=\alpha_{\rm mixup}$.

\medskip
\noindent
{\bf Random erasing with zero value.} We also adopt the {\it Random Erasing} \cite{zhang2017}.
We put zero value on the erasing region of each input instead of a random value as used in the original method.
We set the erasing probability $p=0.5$, the erasing area ratio $S_e\in[0.02,0.25]$, and the erasing aspect ratio $r_e\in[0.3,1]$.
We randomly switch the size of the erasing area from $(H_e,W_e)$ to $(W_e,H_e)$.
%}}}

\subsection{Warmup damping}
\label{subsec:warmup}%{{{
The eigenvalue distribution of the Fisher information matrix (FIM) of deep neural networks
is known to have an extremely long tail \cite{karakida2018},
where most of the eigenvalues are close to zero.
This in turn causes the eigenvalues of the inverse FIM to become extremely large,
which causes the norm of the preconditioned gradient $\mathcal{G}$ to become
huge compared to the parameter $\mathbf{w}$, so the training becomes unstable.
To prevent this problem, we add the {\it damping} \cite{martens2015} value $\gamma$ to the diagonal of the FIM to get a preconditioned gradient:
\begin{equation}
    \mathcal{G}_{\ell}
    =
    \left(\mathbf{F}_{\ell}+\gamma\mathbf{I}\right)^{-1}
    \nabla \mathcal{L}_\ell\,.
\end{equation}
We use a modified {\it Tikhonov damping} technique \cite{martens2015} for a Kronecker-factored FIM (Equation \ref{eq:kf}).
At early stages of the training, the FIM changes rapidly (Figure \ref{fig:cov_change}).
Therefore, we start with a large damping rate and gradually decrease it using following rule:
\begin{align}
    \alpha
    &=
    \frac{
        2\cdot\log_{10}(\gamma^{(0)}/\gamma_{\rm target})
    }{t_{\rm warmup}}\,,
    \\
    \gamma^{(t+1)}
    &=
    (1-\alpha)\gamma^{(t)}
    +
    \alpha\cdot\gamma_{\rm target}\,.
\end{align}

$\gamma^{(t)}$ is the value for the damping in the $t$ th step. 
$\gamma^{(0)}$ is the initial value, and $t_{\rm warmup}>0$ controls the steps to reach
the target value $\gamma_{\rm target}>\gamma^{(0)}$.
At each iteration, we use $\gamma_{\rm BN}^{(t)}=\rho_{\rm BN}\cdot\gamma^{(t)}$ ($\rho_{BN}>1$)
for the Batch Normalization layers to stabilize the training.
%}}}

\subsection{Learning rate and momentum}
\label{subsec:lr}%{{{
The learning rate used for all of our experiments is schedule by {\it polynomial decay}.
The learning rate $\eta^{(e)}$ for $e$ th epoch is determined as follows:
\begin{equation}
    \eta^{(e)}
    =
    \eta^{(0)}
    \cdot
    \left(
        1-
        \frac{e-e_{\rm start}}
        {e_{\rm end} - e_{\rm start}}
    \right)^{p_{\rm decay}}.
\end{equation}
$\eta^{(0)}$ is the initial learning rate and $e_{\rm start},e_{\rm end}$ is the epoch when the decay starts and ends.
The decay rate $p_{\rm decay}$ guides the speed of the learning rate decay.
The learning rate scheduling in our experiments are plotted in Figure \ref{fig:accuracy_epoch}\,.

We use the momentum method for K-FAC updates.
Because the learning rate decays rapidly in the final stage of the training with the polynomial decay, the current update can become smaller than the previous update.
We adjust the momentum rate $m^{(e)}$ for $e$ th epoch so that the ratio between $m^{(e)}$ and $\eta^{(e)}$ is fixed throughout the training: 
\begin{equation}
     m^{(e)}=\frac{m^{(0)}}{\eta^{(0)}}\cdot\eta^{(e)}\,,
\end{equation}
where $m^{(0)}$ is the initial momentum rate.
The weights are updated as follows:
\begin{equation}
    \mathbf{w}^{(t+1)}
    \leftarrow
    \mathbf{w}^{(t)}
    -\eta^{(e)}\,\mathcal{G}^{(t)}
    +m^{(e)}(\mathbf{w}^{(t)}-\mathbf{w}^{(t-1)})\,.
    \label{eq:paramupdate}
\end{equation}
%}}}

\subsection{Weights rescaling}
\label{subsec:rescale}%{{{
To prevent the scale of weights from becoming too large,
we adopt the {\it Normalizing Weights} \cite{vanlaarhoven2017} technique.
We rescale the $\mathbf{w}$ to have a norm $\sqrt{2\cdot d_{\rm out}}$ after (\ref{eq:paramupdate}):
\begin{align}
    \mathbf{w}^{(t+1)}
    &\leftarrow
    \sqrt{2\cdot d_{\rm out}}\cdot
    \frac{
        \mathbf{w}^{(t+1)}
    }
    {
        \|\mathbf{w}^{(t+1)}\|+\epsilon
    }\,.
\end{align}
where we use $\epsilon=1\cdot 10^{-9}$ to stabilize the computation.
$d_{\rm out}$ is the output dimension or channels of the layer.%}}}

\section{Results}
We train ResNet-50 \cite{he2015a} for ImageNet \cite{deng2012} in all of our experiments.
We use the same hyperparameters for the same mini-batch size
when comparing the different schemes in Section \ref{sec:training}.
The training curves for the top-1 validation accuracy shown in
Figures~\ref{fig:accuracy_epoch}, \ref{fig:stale_diag_compare}
are averaged over 2 or 3 executions using the same hyperparameters.
The hyperparameters for our results are shown in Table~\ref{tab:settings}. 
We implement all computation on top of Chainer \cite{akiba2017a,tokui2015}.
%(our Chainer extenstion is available at https://github.com/tyohei/chainerkfac).
We initialize the weights by the {\it HeNormal} initializer of Chainer \footnote{https://docs.chainer.org/en/stable/reference/generated/\\chainer.initializers.HeNormal.html} with the default parameters.

\begin{table*}[ht]%{{{
    \begin{center}
        \caption{Hyperparameters of the training with large mini-batch size (BS) used for our schemes in Section \ref{sec:training}}
        \label{tab:settings}
        \small
        \begin{tabular}{c c c c c c c c c c c}
            \toprule
            \multirow{2}[3]{*}{
            \begin{tabular}{c}
                {\bf BS}
            \end{tabular} 
            } & 
            \begin{tabular}{c}
                {\bf Running} \\
                {\bf mixup} (Sec. \ref{subsec:data}) 
            \end{tabular} 
            & \multicolumn{4}{c}{{\bf Warmup damping} (Sec. \ref{subsec:warmup})} & \multicolumn{5}{c}{{\bf Learning rate and momentum} (Sec. \ref{subsec:lr})} \\
            \cmidrule(lr){2-2}
            \cmidrule(lr){3-6}
            \cmidrule(lr){7-11}
            & $\alpha_{\rm mixup}$ & $\gamma^{(0)}$ & $\gamma_{\rm target}$ & $\rho_{\rm BN}$ & $t_{\rm warmup}$ & $p_{\rm decay}$ & $e_{\rm start}$ & $e_{\rm end}$ & $\eta^{(0)}$ & $m^{(0)}$ \\
            \midrule
            4,096 & $0.4$ & $2.5\cdot10^{-2}$ & $2.5\cdot10^{-4}$ & $16.0$ & $313$ & $11.0$ & $1$ & $53$ & $8.18\cdot10^{-3}$ & $0.997$\\
            8,192 & $0.4$ & $2.5\cdot10^{-2}$ & $2.5\cdot10^{-4}$ & $16.0$ & $157$ & $8.0$ & $1$ & $53.5$ & $1.25\cdot10^{-2}$ & $0.993$\\
            16,384 & $0.4$ & $2.5\cdot10^{-2}$ & $2.5\cdot10^{-4}$ & $32.0$ & $79$ & $8.0$ & $1$ & $53.5$ & $2.5\cdot10^{-2}$ & $0.985$\\
            32,768 & $0.6$ & $2.0\cdot10^{-2}$ & $2.0\cdot10^{-4}$ & $16.0$ & $59$ & $3.5$ & $1.5$ & $49.5$ & $3.0\cdot10^{-2}$ & $0.97$\\
            65,536 & $0.6$ & $1.5\cdot10^{-2}$ & $1.5\cdot10^{-4}$ & $16.0$ & $40$ & $2.9$ & $2$ & $64.5$ & $4.0\cdot10^{-2}$ & $0.95$\\
            131,072 & $1.0$ & $1.0\cdot10^{-2}$ & $1.0\cdot10^{-4}$ & $8.0$ & $30$ & $2.9$ & $3$ & $107.6$ & $7.0\cdot10^{-2}$ & $0.93$\\
            \bottomrule
        \end{tabular}
    \end{center}
\end{table*}%}}}

\subsection{Experiment environment}
We conduct all experiments on the ABCI (AI Bridging Cloud Infrastructure)
\footnote{https://abci.ai/}
operated by the National Institute of Advanced Industrial Science and Technology (AIST) in Japan.
ABCI has 1088 nodes with four NVIDIA Tesla V100 GPUs per node.
Due to the additional memory required by K-FAC, all of our experiments use a mini-batch size of 32 images per GPU.
For large mini-batch size experiments which cannot be executed directly,
we used an accumulation method to mimic the behavior by accumulating over multiple steps.
We were only given a 24 hour window to use the full machine so we had to tune the hyperparameters
on a smaller number of nodes while mimicking the global mini-batch size of the full node run.

\subsection{Scalability}
\begin{figure}[h]
\includegraphics[width={0.98\linewidth}]{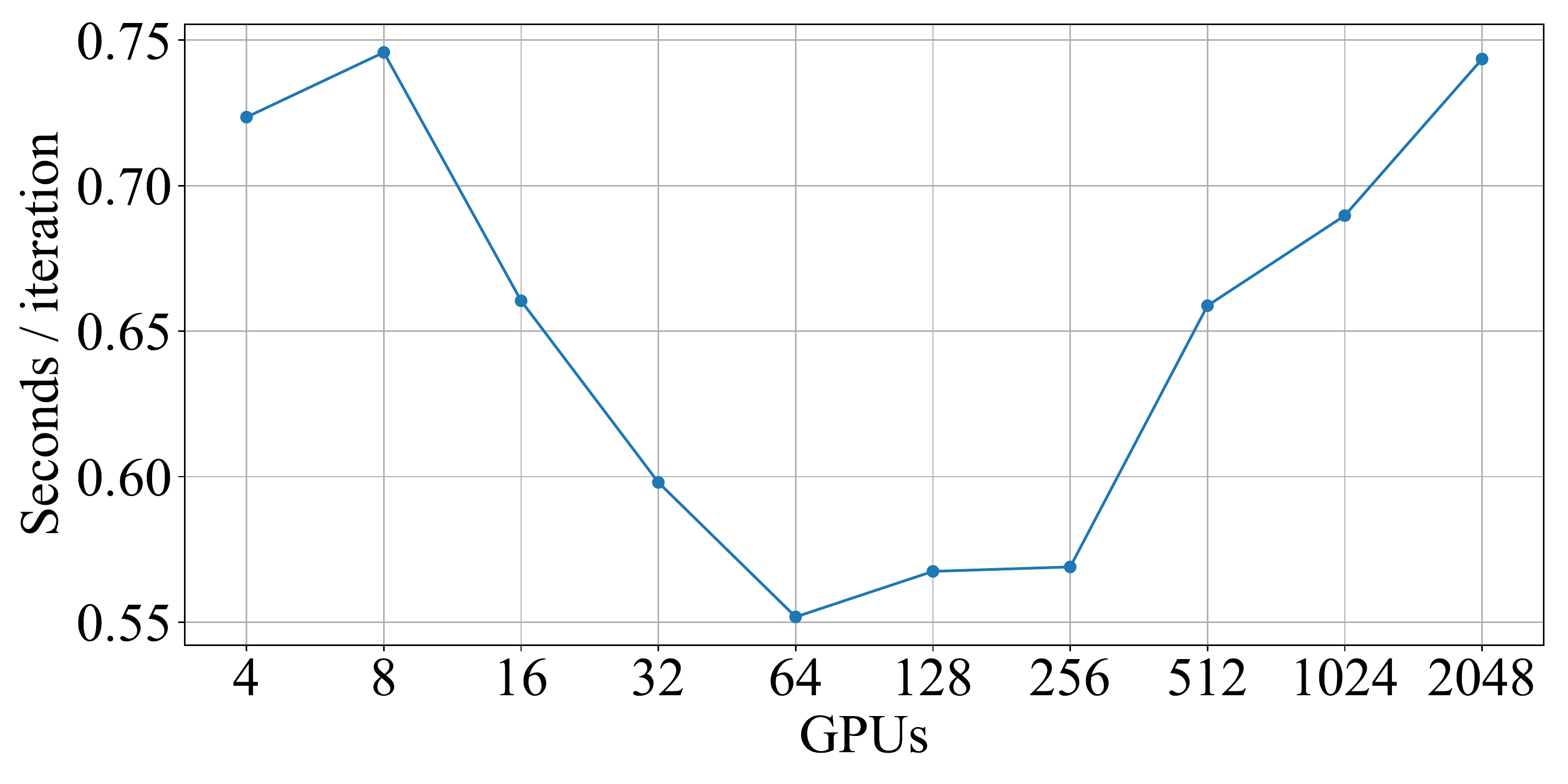}
\caption{\label{fig:iterpsec} Time per iteration of K-FAC on ResNet-50 with ImageNet
using different number of GPUs.
}

\end{figure}

We measured the scalability of our distributed K-FAC implementation on ResNet-50 with ImageNet dataset.
Figure~\ref{fig:iterpsec} shows the time for one iteration using different number of GPUs.
Ideally, this plot should show a flat line parallel to the x-axis,
since we expect the time per iteration to be independent of the number of GPUs.
From 1 GPU to 64 GPUs, we observed a superlinear scaling,
where the 64 GPU case is 131.1\% faster compared to 1 GPU,
which is the consequence of our hybrid data/model-parallel design.
ResNet-50 has 107 layers in total when all the convolution, fully-connected,
and batch normalization layers are accounted for.
Despite this superlinear scaling, after 256 GPUs we observe performance
degradation due to the communication overhead.

\subsection{Large mini-batch training with K-FAC}
\label{subsec:training}
\begin{figure}[h]
    \includegraphics[width={\linewidth}]{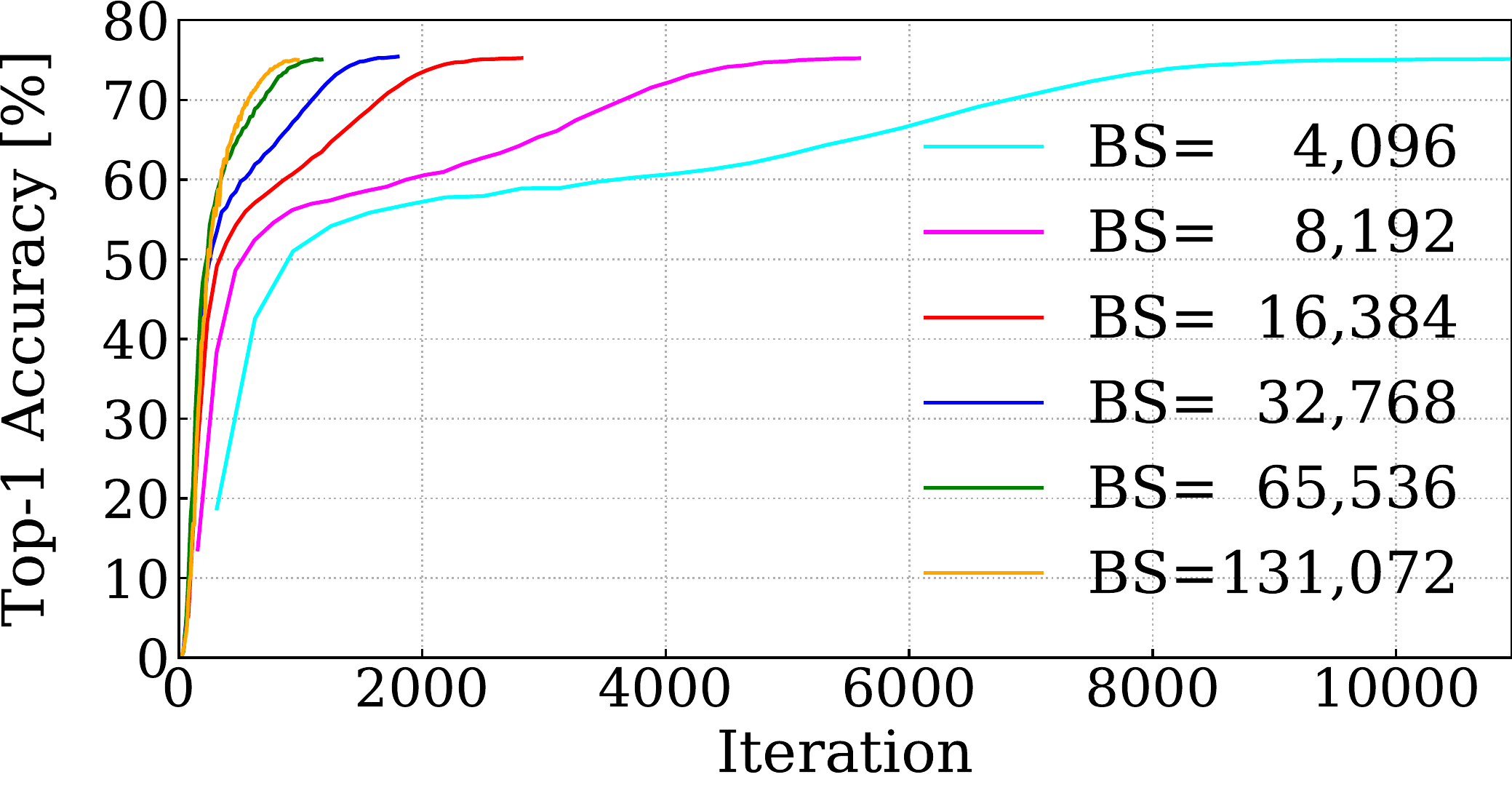}
    \includegraphics[width={\linewidth}]{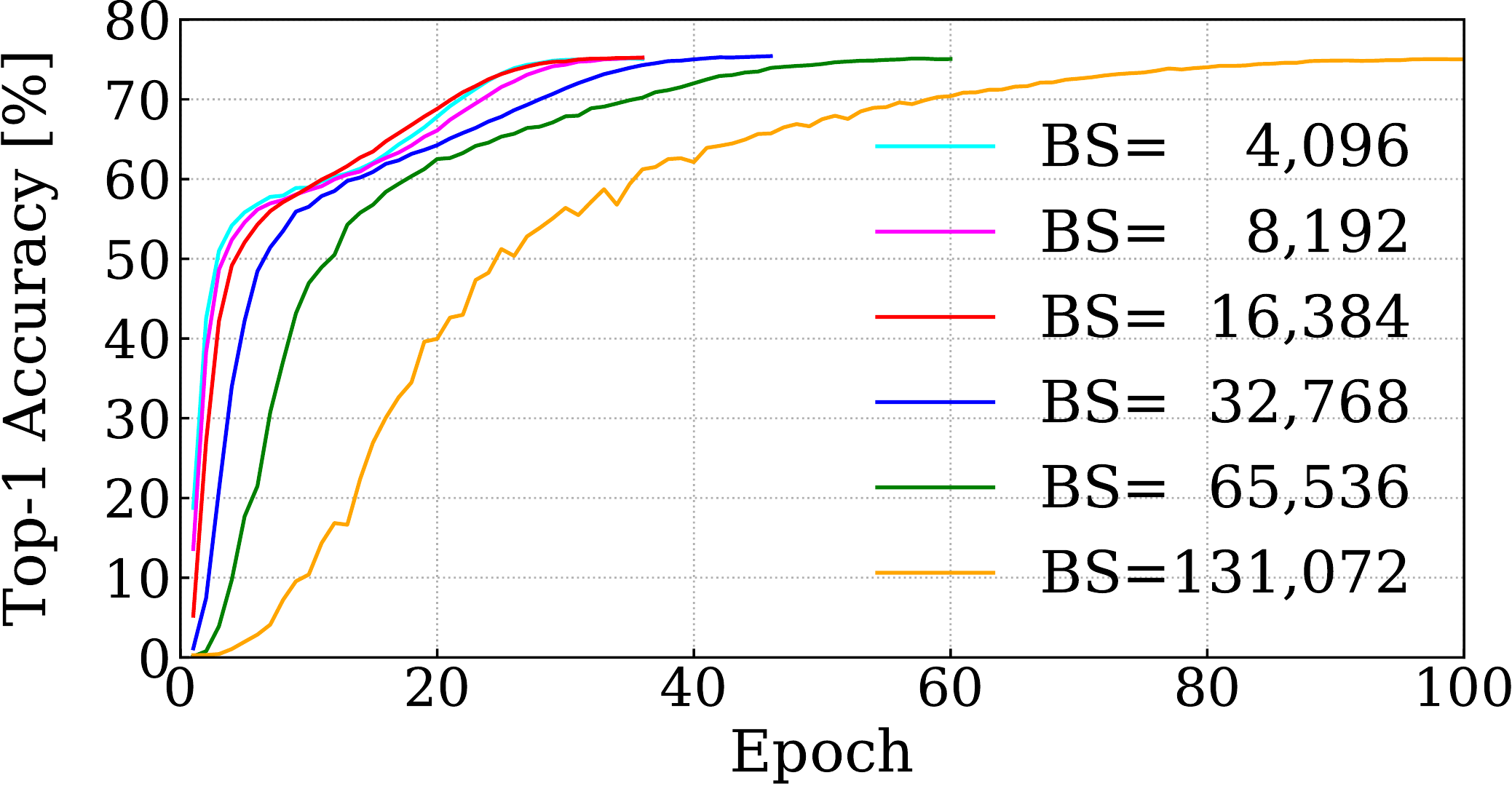}
    \includegraphics[width={\linewidth}]{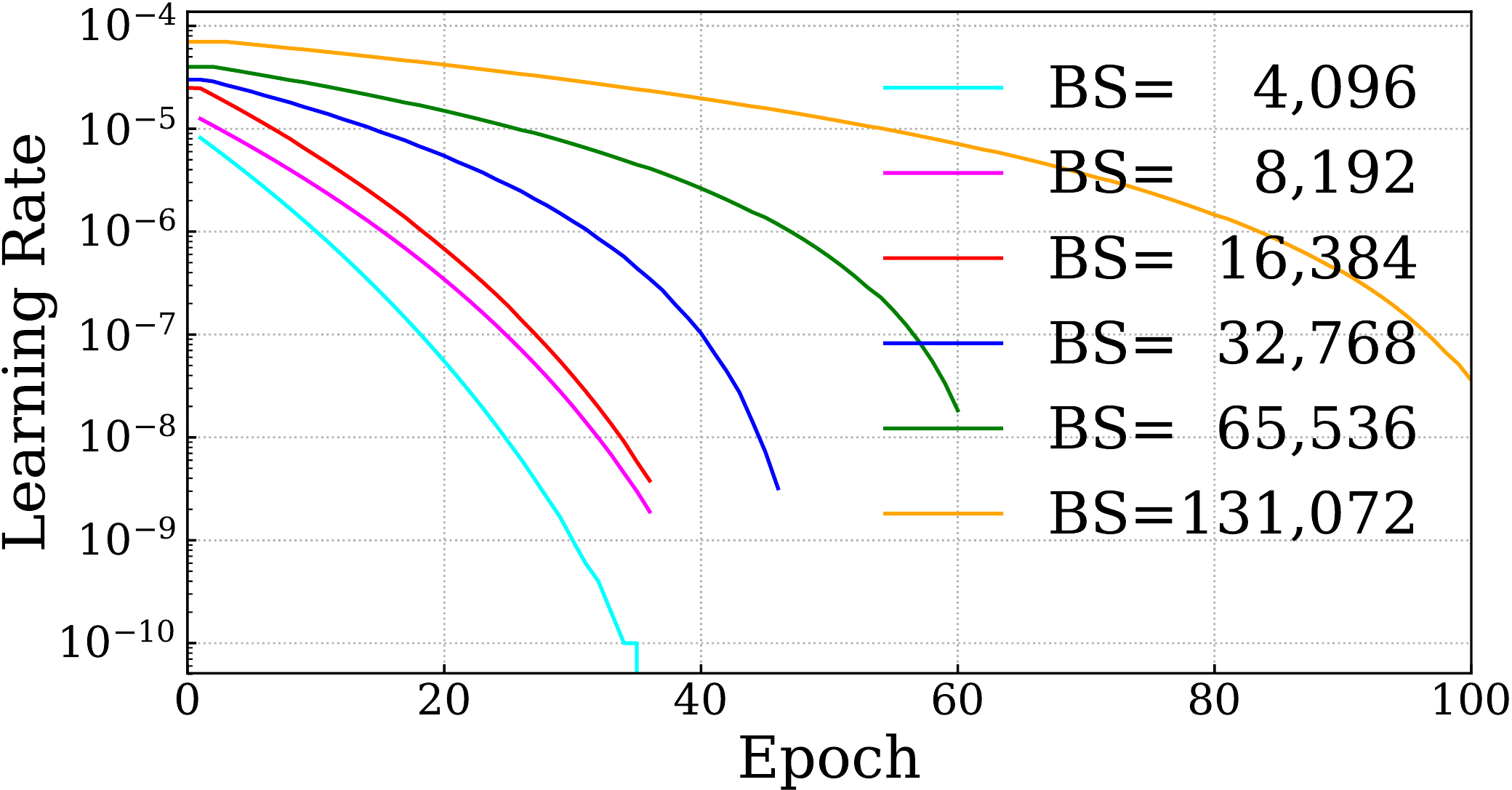}
    \caption{Top-1 validation accuracy and learning rate schedules of training of ResNet-50 for ImageNet with K-FAC}
    \label{fig:accuracy_epoch}
\end{figure}

We trained ResNet-50 for the classification task on ImageNet with extremely large mini-batch size BS=\{4,096 (4K), 8,192 (8K), 16,384 (16K), 32,768 (32K), 65,536 (65K), 131,072 (131K)\} and achieved a competitive top-1 validation accuracy ($\geq75\%$). 
The summary of the training is shown in Table~\ref{tab:summary}.
The training curves and the learning rate schedules are plotted in Figure~\ref{fig:accuracy_epoch}.
When we use BS=\{4K, 8K, 16K, 32K, 65K\}, the training converges in much less than 90 epochs,
which is the usual number of epochs required by SGD-based training of ImageNet \cite{akiba2017,goyal2017,jia2018,mikami2018,you2017}. 
For BS=\{4K,8K,16K\}, the required epochs to reach higher than 75\% top-1 validation accuracy does not change so much.
Even for a relatively large mini-batch size of BS=32K, K-FAC still converges
in half the number of epochs compared to SGD.
When increasing the mini-batch size to BS=65K, we see a 33\% increase in the number of epochs it takes to converge.
Note that the calculation time is still decreasing while the number of epochs
is less than double when we double the mini-batch size,
assuming that doubling the mini-batch corresponds to doubling the number of GPUs (and halving the execution time).
At BS=131K, there are less than 10 iterations per epoch since the dataset size of ImageNet is 1,281,167.
None of the SGD-based training of ImageNet have sustained the top-1 validation accuracy at this mini-batch size.
Furthermore, this is the first work that uses K-FAC for the training with extremely large mini-batch size
BS=\{16K,32K,65K,131K\} and achieves a competitive top-1 validation accuracy.

\subsection{Fisher information and large mini-batch training}
\label{subsec:fisher}
We analyzed the relationship between the large mini-batch training with K-FAC
and the Fisher information matrix of ResNet-50.

\begin{figure}[h]
\includegraphics[width={\linewidth}]{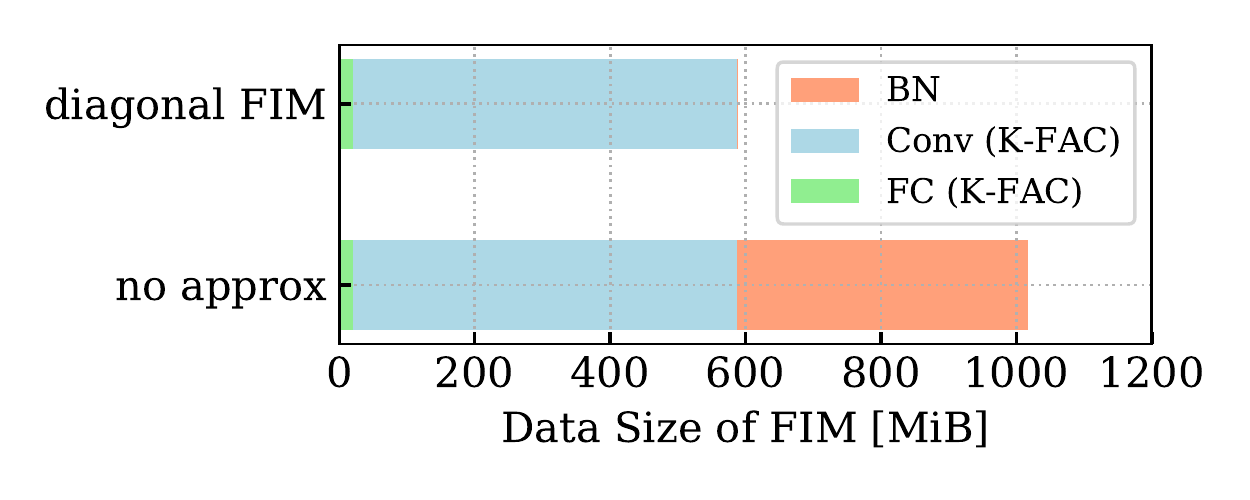}
\caption{Data size of FIM (Kronecker factors) of all layers of ResNet-50 for ImageNet classification per GPU. {\bf no approx}: K-FAC using FIM for Batch Normalization (BN) layers without any approximation. {\bf diagonal FIM}: K-FAC using diagonal FIM for BN layers.} 
\label{fig:cov_nbytes}
\end{figure}

\begin{figure}[h]
\includegraphics[width={0.98\linewidth}]{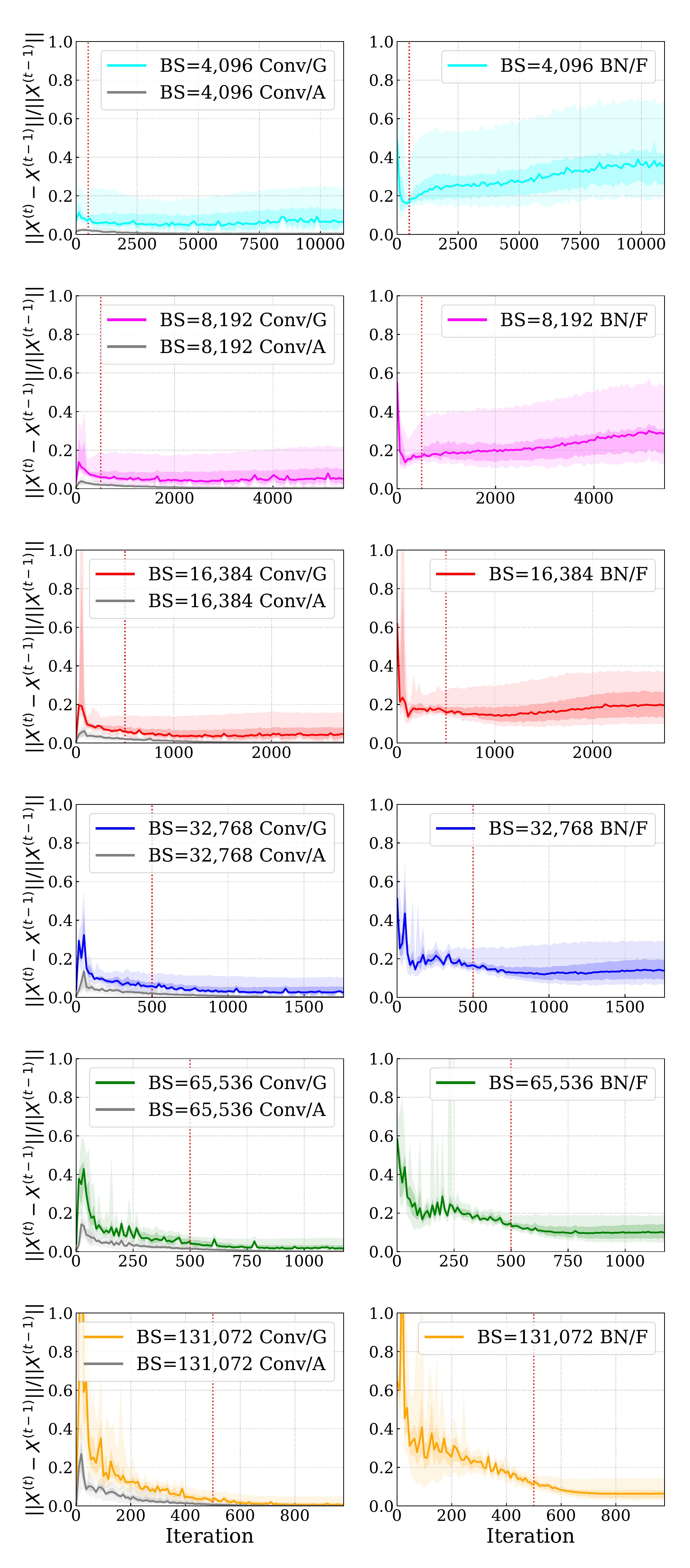}
\caption{Change in the value of the Kronecker Factors ($\mathbf{A},\mathbf{G}$) for convolutional layers (left) and the FIM ($\mathbf{F}$) for Batch Normalization (BN) layers (right) in our large mini-batch training (\ref{subsec:training}).
Each plot shows $\{5,25,50,75,95\}$th percentile of the value among all layers in ResNet-50.
The red line in each plot shows the 500 th iteration.
}
\label{fig:cov_change}
\end{figure}

\medskip
\noindent
%{{{
{\bf Staleness of Fisher information.}
To achieve faster training with distributed K-FAC, reducing the computation and the communication
of the FIM (or the Kronecker factors) is required.   
In ResNet-50 for ImageNet classification, the data of the Kronecker factors $\mathbf{A},\mathbf{G}$
for the convolutional layers and the FIM $\mathbf{F}$ for the Batch Normalization layers
are dominant (Figure~\ref{fig:cov_nbytes}).
Note that we do not factorize the FIM for the Batch Normalization layers into $\mathbf{A}$ and $\mathbf{G}$.
Previous work on K-FAC used stale Kronecker factors by only calculating them every few steps \cite{martens2015}.
Even though our efficient distributed scheme minimizes the overhead of the Kronecker factor calculation,
we thought it was worth investigating how much staleness we can tolerate to further speed up our method.
We examine the change rate of the Kronecker factors for the convolutional layers
and the FIM for the Batch Normalization layers.
\begin{equation}
    {\rm Diff}^{(t)}=
    \frac{
    \|X^{(t)}-X^{(t-1)}\|_{F}
    }{
    \|X^{(t-1)}\|_{F}
    }\,.
\end{equation}
where $\|\cdot\|_{F}$ is the {\it Frobenius} norm.
The results from our large mini-batch training in Section \ref{subsec:training} are plotted in Figure~\ref{fig:cov_change}.
We can see that the FIM fluctuates less for larger mini-batches,
because each mini-batch becomes more statistically stable.
This implies that we can reduce the frequency of updating the FIM more aggressively for larger mini-batches.
For the convolutional layers, the Kronecker factor $\mathbf{A}_{\ell-1}$ which represents
the correlation among the dimensions of the input of the $\ell$ th layer ($\ell=1,\dots,L$)
fluctuates less than $\mathbf{G}_\ell$ which represents the correlation among the dimensions
of the gradient for the output of the $\ell$ th layer.
Hence, we can also consider refreshing $\mathbf{A}_{\ell-1}$ less frequently than $\mathbf{G}_\ell$ .%}}}

\medskip
\noindent%{{{
{\bf Training with stale Fisher information.}
We found that, regardless of the mini-batch size,
the FIM changes rapidly during the first 500 or so iterations.
Based on this, we reduce the frequency of updating ($\mathbf{A},\mathbf{G},\mathbf{F}$)
after 500 iterations.
We apply a heuristic scheduling of the refreshing interval.
The refreshing interval (iterations) ${\rm interval}^{(e)}$ for the $e$ th epoch is determined by: 
\begin{equation}
    {\rm interval}^{(e)}
    =
    \min(20,5\cdot\lfloor e/5\rfloor+1)\,.
\end{equation}
Using 1024 NVIDIA Tesla V100, we achieve 74.9 \% top-1 accuracy with ResNet-50 for ImageNet
in 10 minutes (45 epochs, including a validation after each epoch).
We used the same hyperparameters shown in Table~\ref{tab:settings}.
The training time and the validation accuracy are competitive with the results reported
by related work that use SGD for training (the comparison is shown in Table~\ref{tab:compare}).

%}}}

\medskip
\noindent%{{{
{\bf Diagonal Fisher information matrix.}
\begin{figure}[ht]
\includegraphics[width={\linewidth}]{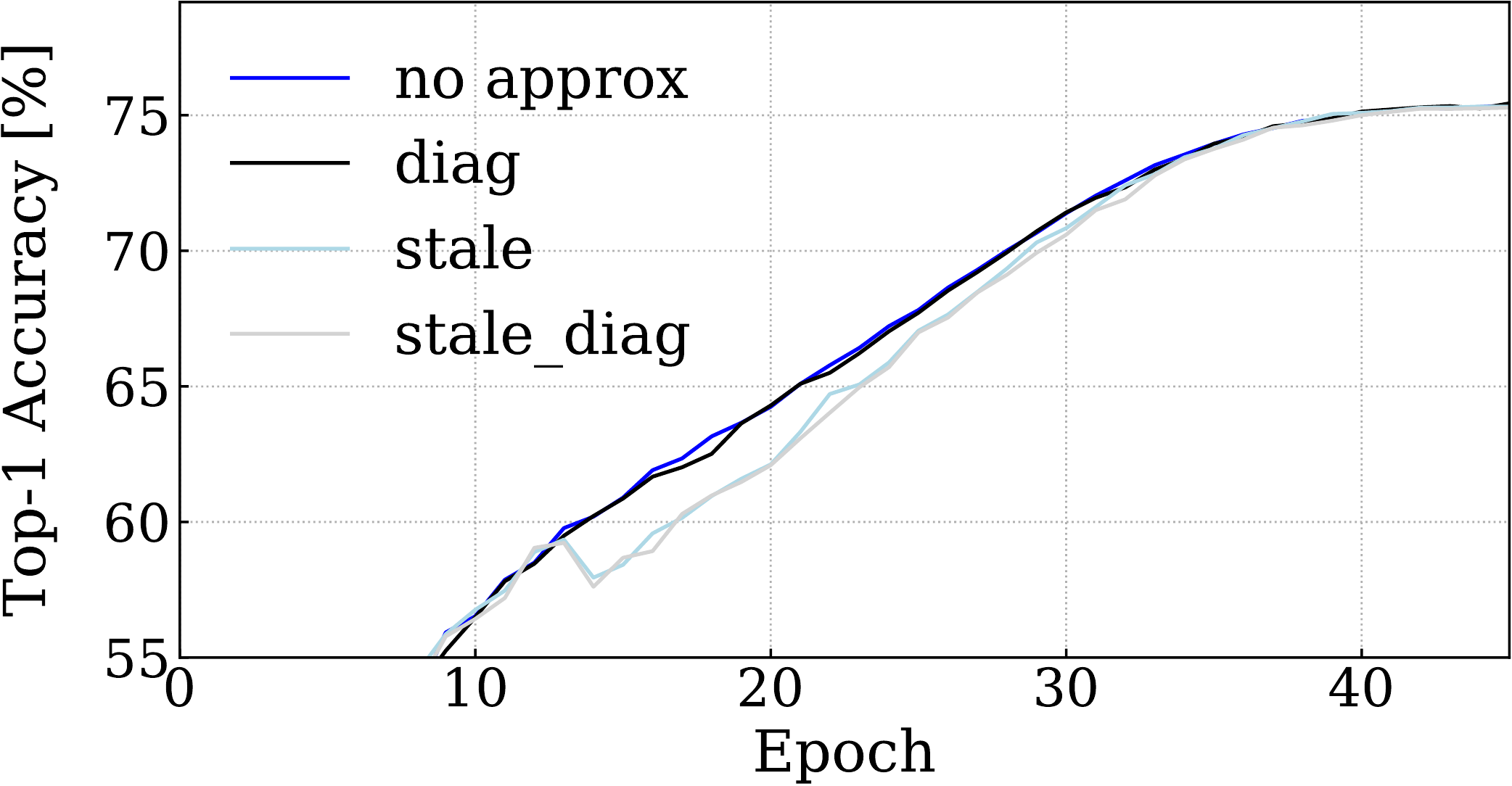}
\caption{Top-1 validation accuracy of training of ResNet-50 for ImageNet with K-FAC using BS=32,768.
    {\bf no approx}: the same training curve with K-FAC plotted in Figure~\ref{fig:accuracy_epoch}.
    {\bf diag}: diagonal FIM for all Batch Normalization layers.
    {\bf stale}: stale FIM for all layers.
    {\bf stale\_diag}: apply both $\rm diag$ and $\rm stale$.
}
\label{fig:stale_diag_compare}
\end{figure}
As shown in Figure~\ref{fig:cov_nbytes}, the FIM for Batch Normalization (BN) layers contribute to
a large portion of the memory overhead of K-FAC.
To alleviate this overhead, we approximate it with a diagonal matrix.
By using the diagonal approximation, we can reduce the memory consumption of the FIM
for the all layers of ResNet-50
from 1017MiB to 587MiB (Figure~\ref{fig:cov_nbytes}).
We measure the effect of the diagonal approximation on the accuracy of ResNet-50 for ImageNet
with mini-batch size BS=32,768 with/without using stale Fisher information for all layers. 
In this experiment, we adopt another heuristic for ${\rm interval}^{(e)}$:
\begin{equation}
    \begin{aligned}
        {\rm interval}^{(e)}
        =
        \left\{
            \begin{aligned}
                &1&&\mbox{ if  }e<13\,, \\
                &20&&\mbox{ otherwise}\,.
            \end{aligned}
        \right.
    \end{aligned}
\end{equation}
The training curves are plotted in Figure~\ref{fig:stale_diag_compare}.
Using diagonal FIM does not affect the training curve even with stale FIM.
This result suggests that only diagonal values of the FIM is essential for the training of BN layers.%}}}

\section{Conclusion}
In this work, we proposed a large-scale distributed computational design for the second-order optimization
using Kronecker-Factored Approximate Curvature (K-FAC) and showed the advantages of K-FAC
over the first-order stochastic gradient descent (SGD) for the training of ResNet-50 with ImageNet classification
using extremely large mini-batches.
We introduced several schemes for the training using K-FAC with mini-batch sizes up to 131,072
and achieved over 75\% top-1 accuracy in much fewer number of epochs/iterations compared to
the existing work using SGD with large mini-batch.
Contrary to prior claims that second order methods do not generalize as well as SGD,
we were able to show that this is not at all the case, even for extremely large mini-batches.
Data and model hybrid parallelism introduced in our design allowed us to train on 1024 GPUs
and achieved 74.9\% in 10 minutes by using K-FAC with the stale Fisher information matrix (FIM).
This is the first work which observes the relationship between the FIM of ResNet-50
and its training on large mini-batches ranging from 4K to 131K.
There is still room for improvement in our distributed design to overcome the bottleneck of
computation/communication for K-FAC -- the Kronecker factors can be approximated more aggressively
without loss of accuracy.
One interesting observation is that,
whenever we coupled our method with a well known technique that improves the convergence of SGD,
it allowed us to approximate the FIM more aggressively without any loss of accuracy.
This suggests that all these seemingly \textit{ad hoc} techniques to improve the convergence of SGD,
are actually performing an equivalent role to the FIM in some way.
The advantage that we have in designing better optimizers by taking this approch
is that we are starting from the most mathematically rigorous form,
and every improvement that we make is a systematic design decision based on observation of the FIM.
Even if we end up having similar performance to the best known first-order methods,
at least we will have a better understanding of why it works by starting from second-order methods.
Further analysis of the eigenvalues of FIM and its effect on preconditioning the gradient
will allow us to further understand the advantage of second-order methods
for the training of deep neural networks with extremely large mini-batches.

\section*{Acknowledgements}
Computational resource of AI Bridging Cloud Infrastructure (ABCI) was awarded by "ABCI Grand Challenge" Program,
National Institute of Advanced Industrial Science and Technology (AIST).
This work is supported by JST CREST Grant Number JPMJCR19F5, Japan.
This work was supported by JSPS KAKENHI Grant Number JP18H03248.
(Part of) This work is conducted as research activities of AIST - Tokyo Tech Real World Big-Data Computation
Open Innovation Laboratory (RWBC-OIL).
This work is supported by "Joint Usage/Research Center for Interdisciplinary Large-scale
Information Infrastructures" in Japan (Project ID: jh180012-NAHI).

{\small
\bibliographystyle{ieee}
\bibliography{egbib}

\begin{thebibliography}{10}\itemsep=-1pt

\bibitem{martin2015}
M.~Abadi, P.~Barham, J.~Chen, Z.~Chen, A.~Davis, J.~Dean, M.~Devin,
  S.~Ghemawat, G.~Irving, M.~Isard, M.~Kudlur, J.~Levenberg, R.~Monga,
  S.~Moore, D.~G. Murray, B.~Steiner, P.~Tucker, V.~Vasudevan, P.~Warden,
  M.~Wicke, Y.~Yu, and X.~Zheng.
\newblock {TensorFlow}: A system for large-scale machine learning.
\newblock In {\em OSDI}, pages 265--283, 2016.

\bibitem{akiba2017a}
T.~Akiba, K.~Fukuda, and S.~Suzuki.
\newblock {ChainerMN}: Scalable distributed deep learning framework.
\newblock In {\em Workshop on Machine Learning Systems in NIPS}, 2017.

\bibitem{akiba2017}
T.~Akiba, S.~Suzuki, and K.~Fukuda.
\newblock Extremely large minibatch sgd: Training {ResNet-50} on {ImageNet} in
  15 minutes.
\newblock {\em arXiv preprint arXiv:1711.04325}, 2017.

\bibitem{amari1998}
S.-I. Amari.
\newblock Natural gradient works efficiently in learning.
\newblock {\em Neural Computation}, 10(2):251--276, 1998.

\bibitem{ba2017}
J.~Ba, R.~Grosse, and J.~Martens.
\newblock Distributed second-order optimization using {Kronecker}-factored
  approximations.
\newblock In {\em ICLR}, 2017.

\bibitem{botev2017}
A.~Botev, H.~Ritter, and D.~Barber.
\newblock Practical {{Gauss}}-{Newton} optimisation for deep learning.
\newblock In {\em ICML}, pages 557--565, 2017.

\bibitem{deng2012}
J.~Deng, W.~Dong, R.~Socher, L.-J. Li, K.~Li, and L.~{Fei-Fei}.
\newblock {{ImageNet}}: A large-scale hierarchical image database.
\newblock In {\em CVPR}, pages 248--255, 2009.

\bibitem{devarakonda2017}
A.~Devarakonda, M.~Naumov, and M.~Garland.
\newblock {AdaBatch}: Adaptive batch sizes for training deep neural networks.
\newblock {\em arXiv preprint arXiv:1712.02029}, 2017.

\bibitem{goyal2017}
P.~Goyal, P.~Dollar, R.~Girshick, P.~Noordhuis, L.~Wesolowski, A.~Kyrola,
  A.~Tulloch, Y.~Jia, and K.~He.
\newblock Accurate, large minibatch {SGD}: Training {ImageNet} in 1 hour.
\newblock {\em arXiv preprint arXiv:1706.02677}, 2017.

\bibitem{grosse2016}
R.~Grosse and J.~Martens.
\newblock A {Kronecker}-factored approximate {Fisher} matrix for convolution
  layers.
\newblock In {\em ICML}, pages 573--582, 2016.

\bibitem{guo2018}
H.~Guo, Y.~Mao, and R.~Zhang.
\newblock {{MixUp}} as locally linear out-of-manifold regularization.
\newblock {\em arXiv preprint arXiv:1809.02499}, 2018.

\bibitem{he2015a}
K.~He, X.~Zhang, S.~Ren, and J.~Sun.
\newblock Deep residual learning for image recognition.
\newblock In {\em CVPR}, pages 770--778, 2016.

\bibitem{hoffer2018}
E.~Hoffer, I.~Hubara, and D.~Soudry.
\newblock Train longer, generalize better: Closing the generalization gap in
  large batch training of neural networks.
\newblock In {\em NIPS}, pages 1731--1741, 2017.

\bibitem{ioffe2015}
S.~Ioffe and C.~Szegedy.
\newblock {Batch Normalization}: Accelerating deep network training by reducing
  internal covariate shift.
\newblock In {\em ICML}, pages 448--456, 2015.

\bibitem{jia2018}
X.~Jia, S.~Song, W.~He, Y.~Wang, H.~Rong, F.~Zhou, L.~Xie, Z.~Guo, Y.~Yang,
  L.~Yu, T.~Chen, G.~Hu, S.~Shi, and X.~Chu.
\newblock Highly scalable deep learning training system with mixed-precision:
  Training {ImageNet} in four minutes.
\newblock {\em arXiv preprint arXiv:1807.11205}, 2018.

\bibitem{karakida2018}
R.~Karakida, S.~Akaho, and S.-i. Amari.
\newblock Universal statistics of {Fisher} information in deep neural networks:
  Mean field approach.
\newblock {\em arXiv preprint arXiv:1806.01316}, 2018.

\bibitem{leroux2008}
N.~Le~Roux, P.-A. Manzagol, and Y.~Bengio.
\newblock Topmoumoute online natural gradient algorithm.
\newblock In {\em NIPS}, pages 849--856, 2008.

\bibitem{lin2018}
T.~Lin, S.~U. Stich, and M.~Jaggi.
\newblock Don't use large mini-batches, use local {SGD}.
\newblock {\em arXiv preprint arXiv:1808.07217}, 2018.

\bibitem{martens2010}
J.~Martens.
\newblock Deep learning via {Hessian}-free optimization.
\newblock In {\em ICML}, pages 735--742, 2010.

\bibitem{martens2018}
J.~Martens, J.~Ba, and M.~Johnson.
\newblock Konecker-factored curvature approximations for recurrent neural
  networks.
\newblock In {\em ICLR}, 2018.

\bibitem{martens2015}
J.~Martens and R.~Grosse.
\newblock Optimizing neural networks with {Kronecker}-factored approximate
  curvature.
\newblock In {\em ICML}, pages 2408--2417, 2015.

\bibitem{mikami2018}
H.~Mikami, H.~Suganuma, P.~{U-chupala}, Y.~Tanaka, and Y.~Kageyama.
\newblock Massively distributed {SGD}: {ImageNet/ResNet-50} training in a
  flash.
\newblock {\em arXiv preprint arXiv:1811.05233}, 2018.

\bibitem{shallue2018}
C.~J. Shallue, J.~Lee, J.~Antognini, J.~Sohl-Dickstein, R.~Frostig, and G.~E.
  Dahl.
\newblock Measuring the effects of data parallelism on neural network training.
\newblock {\em arXiv preprint arXiv:1811.03600}, 2018.

\bibitem{smith2017}
S.~L. Smith, P.-J. Kindermans, and Q.~V. Le.
\newblock Don't decay the learning rate, increase the batch size.
\newblock In {\em ICLR}, 2018.

\bibitem{tokui2015}
S.~Tokui, K.~Oono, S.~Hido, and J.~Clayton.
\newblock {Chainer}: a next-generation open source framework for deep learning.
\newblock In {\em Workshop on Machine Learning Systems in NIPS}, 2015.

\bibitem{vanlaarhoven2017}
T.~{van Laarhoven}.
\newblock L2 regularization versus batch and weight normalization.
\newblock {\em arXiv preprint arXiv:1706.05350}, 2017.

\bibitem{wu2017}
Y.~Wu, E.~Mansimov, S.~Liao, R.~Grosse, and J.~Ba.
\newblock Scalable trust-region method for deep reinforcement learning using
  {Kronecker}-factored approximation.
\newblock In {\em NIPS}, pages 5279--5288, 2017.

\bibitem{ying2018a}
C.~Ying, S.~Kumar, D.~Chen, T.~Wang, and Y.~Cheng.
\newblock Image classification at supercomputer scale.
\newblock In {\em Workshop on Systems for ML in NIPS}.

\bibitem{you2017}
Y.~You, Z.~Zhang, C.-J. Hsieh, J.~Demmel, and K.~Keutzer.
\newblock {ImageNet} training in minutes.
\newblock In {\em ICPP}, 2018.

\bibitem{zhang2017}
G.~Zhang, S.~Sun, D.~Duvenaud, and R.~Grosse.
\newblock Noisy natural gradient as variational inference.
\newblock In {\em ICML}, pages 5847--5856, 2018.

\bibitem{zhang2017a}
H.~Zhang, M.~Cisse, Y.~N. Dauphin, and D.~{Lopez-Paz}.
\newblock Mixup: Beyond empirical risk minimization.
\newblock In {\em ICLR}, 2018.

\end{thebibliography}
}

\end{document}